\pdfoutput=1
\documentclass[11pt]{article}

\usepackage{acl}
\usepackage{times}
\usepackage{latexsym}

\usepackage[T1]{fontenc}
\usepackage[utf8]{inputenc}

\usepackage{microtype}

\usepackage{inconsolata}

\usepackage{graphicx}

\usepackage{microtype}
\usepackage{hyperref}
\usepackage{url}
\usepackage{booktabs}

\usepackage{lineno}

\definecolor{darkblue}{rgb}{0, 0, 0.5}
\hypersetup{colorlinks=true, citecolor=darkblue, linkcolor=darkblue, urlcolor=darkblue}

\usepackage{graphicx}
\usepackage{adjustbox}
\usepackage{amssymb}
\usepackage{amsmath} 
\usepackage{mathtools}
\usepackage{latexsym}
\usepackage{booktabs}
\usepackage{subcaption}
\usepackage[T1]{fontenc}
\usepackage[linesnumbered,ruled,vlined]{algorithm2e}
\newcommand{\ourmethod}{ProxyReward\xspace}
\newcommand{\task}{Open-LTG\xspace}

\usepackage{colortbl}
\usepackage{multirow}

\usepackage{tcolorbox}
\definecolor{lightgreen}{RGB}{145, 204, 117}
\definecolor{lightyellow}{RGB}{250, 200, 88}
\definecolor{lightred}{RGB}{238, 102, 102}
\definecolor{lightblue}{RGB}{115, 192, 222}
\newtcolorbox{promptbox}[2][Prompt]{
  colback=black!5!white,
  arc=5pt, 
  boxrule=0.5pt,
  fonttitle=\bfseries,
  title=#1, 
  before upper={\small}, % 修改为 \small 以设置为 10pt
  fontupper=\fontfamily{ptm}\selectfont, 
  colframe=#2, % 使用传递的参数来设定 colframe
}

\title{From General to Targeted Rewards: Surpassing GPT-4 in Open-Ended Long-Context Generation}

\author{
    \textbf{Zhihan~Guo$^1$},
    \textbf{Jiele~Wu$^2$},
    \textbf{Wenqian~Cui$^1$},
    \textbf{Yifei~Zhang$^3$},
    \textbf{Minda~Hu$^1$},
    \textbf{Yufei~Wang$^4$},
    \textbf{Irwin~King$^{1}$}
\\
 \textsuperscript{1}The Chinese University of Hong Kong, Hong Kong SAR, China
 \\
 \textsuperscript{2}National University of Singapore, Singapore, Singapore
 \\
  \textsuperscript{3}Nanyang Technological University, Singapore, Singapore
 \\
 \textsuperscript{4}Macquarie University, Sydney, Australia
\\
\texttt{\{zhguo22,king\}@cse.cuhk.edu.hk} 
}
\begin{document}
\maketitle
\begin{abstract}
Current research on long-form context in Large Language Models (LLMs) primarily focuses on the understanding of long-contexts, the Open-ended Long Text Generation (\task) remains insufficiently explored. Training a long-context generation model requires curation of gold-standard reference data, which is typically nonexistent for informative \task tasks. However, previous methods only utilize general assessments as reward signals, which limits accuracy. To bridge this gap, we introduce \textbf{\ourmethod}, an innovative reinforcement learning (RL) based framework, which includes a dataset and a reward signal computation method. Firstly, \textbf{\ourmethod Dataset} generation is accomplished through simple prompts that enables the model to create automatically, obviating extensive labeled data or significant manual effort. Secondly, \textbf{\ourmethod Signal} offers a targeted evaluation of information comprehensiveness and accuracy for specific questions. The experimental results indicate that our method \ourmethod surpasses even GPT-4-Turbo. It can significantly enhance performance by 20\% on the \task task when training widely used open-source models, while also surpassing the LLM-as-a-Judge approach. Our work presents effective methods to enhance the ability of LLMs to address complex open-ended questions posed by humans.
\end{abstract}

\section{Introduction}
\label{Introduction}
\begin{figure}[t!]
    \centering
    \includegraphics[width=1\linewidth]{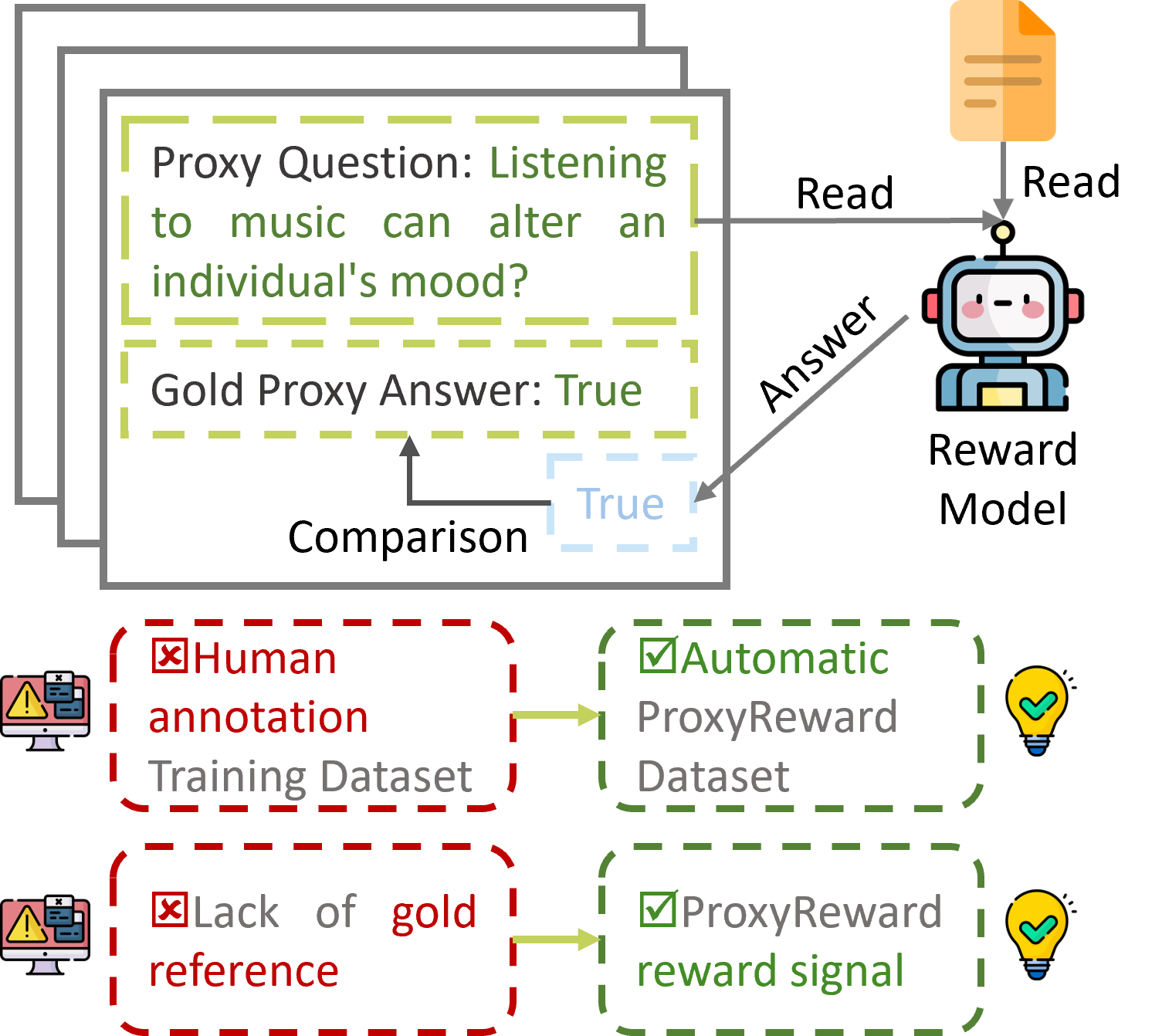}
    \caption{Illustration of the \ourmethod model and its two advantages: automatic over human data label annotation and the existence of reward signals for Open-LTG problems.}
    \label{fig:first}
\end{figure}

Open-ended Long Text Generation (\task) represents a significant challenge in the field of large language model (LLM) research\cite{kumar2024longlamp, lee2022factuality}, owing to its inherent openness and complexity \cite{li2023contrastive, sudhakaran2023mariogpt, brown2020language, touvron2023llama}. While current research has made substantial progress in enhancing LLMs' ability to understand long contexts~\cite{wu2024visual}, the complementary challenge of generating coherent, informative, and contextually grounded outputs spanning thousands of tokens remains critically underexplored \cite{bai2024longwriter, li2024loogle,li2024long, liu2025comprehensive}. As illustrated in Figure~\ref{fig:first}, designing targeted training reward signals represents an essential component in addressing this challenge.

% Open-ended Long Text Generation (\task) represents a significant challenge in the field of large language model (LLM) research, owing to its inherent openness and complexity. However, the complementary challenge of long-form text generation—producing coherent, informative, and contextually grounded outputs spanning thousands of tokens—remains underexplored. While current research has primarily focused on understanding long contexts, the generation aspect remains insufficiently explored. As illustrated in Figure~\ref{fig:first}, mainstream Large Language Models (LLMs) demonstrate limited capabilities in long-context generation. Although models like Deepseek exhibit stronger performance in this domain, their technical details have not been publicly disclosed.

The \task task faces several fundamental obstacles that have impeded progress in this domain~\cite{que2024hellobench, liu2024longgenbench}. Unlike traditional LLM generation tasks which have clear reference answers, \task lacks standard answers due to the inherent complexity of information in long texts~\cite{koksal2023longform}. Consequently, \task responses do not follow a single pattern, resulting in insufficient reward signals during model training \cite{zhang2024longreward}. Moreover, annotation of \task data requires the creation of substantial high-quality long-form content~\cite{pham2024suri}, traditionally demanding domain experts with extensive knowledge, leading to inefficiency and prohibitive costs~\cite{micheletti2024exploration}. Thirdly, experts' subjective preferences regarding long-text quality often result in inconsistent evaluations~\cite{tan2024proxyqa, bai2024longbench}, while the diversity of long-form content further complicates efforts to translate subjective assessments to objective evaluation metrics~\cite{xu2023critical,krishna2023longeval}.

% Current research on Open-ended Long Text Generation (\task) faces numerous challenges. Unlike traditional LLM generation tasks, \task lacks standard answers due to the inherent complexity of information in long texts~\cite{koksal2023longform}. Consequently, \task responses do not follow a single pattern, resulting in insufficient reward signals during model training. Additionally, annotation of \task data requires the creation of substantial high-quality long-form content. Traditional manual annotation methods demand domain experts with extensive knowledge, leading to inefficiency and high costs. Moreover, experts' subjective preferences regarding long-text quality can result in inconsistent evaluations. The diversity of long-form content further complicates efforts to translate subjective assessments into objective evaluation metrics.

To bridge this critical gap, this paper introduces \ourmethod, an innovative reinforcement learning (RL)-based framework designed specifically for open-ended long-context generation. The framework consists of two key components. First, the \ourmethod dataset includes 9,271 long-form meta-questions across multiple domains and associated multiple proxy question-answer pairs generated via simple prompts. This dataset is constructed automatically without the need of extensive labeled data or significant manual effort. Second, the \ourmethod signal transforms the subjective and challenging task of evaluating long-form content quality into a long-context understanding task at which has been fully explored on the long-context LLMs~\cite{li2024long}. This approach effectively provides targeted reward signals for more effective model training.

The central innovation of \ourmethod lies in its unique (meta-question, proxy question-answer pairs) structure. For each meta-question, the proxy question-answer pairs comprise multiple questions and corresponding boolean answers, which similar to reading comprehension tests. Our approach leverages these proxy question-answer pairs to derive informative reward signals for reinforcement learning, eliminating the need for gold standard responses and encouraging the generation of more informative and comprehensive outputs. By transforming subjective evaluation of long-form content as an objective assessment task grounded in the reading comprehension abilities of language models, \ourmethod effectively addresses the core challenges of \task tasks.

Our main contributions are as follows:

\begin{itemize}
    \item The \ourmethod Dataset is constructed through a simple, scalable process without relying on predefined response patterns, eliminating the need for costly and labor-intensive supervised annotation.
    \item The \ourmethod serves as a targeted reward mechanism. It leverages the long-context understanding capabilities of LLMs to generate informative reward signals that guide model optimization without relying on traditional gold references.
    \item Experiments demonstrate that \ourmethod significantly enhances performance on the ProxyQA benchmark (increase 20\%) applied to Qwen and Llama models, and even surpassing GPT-4-Turbo. This validates its effectiveness for the \task task.
\end{itemize}

\section{Related Work}

\subsection{Long-context Language Model}
Long-context language models are designed to overcome the context length limitations of language models, enabling them to support a wide range of long context tasks effectively \cite{xiong2023effective,ma2024megalodon}. A prominent line of research focuses on improving the transformer architecture through efficient attention mechanisms \cite{taylong,tay2022efficient,zaheer2020big,jiangminference}, structured state space models \cite{guefficiently,poli2023hyena} or memory recurrent \cite{bulatov2022recurrent}, in order to alleviate the context length limitations. However, these approaches often involve significant approximations that deviate from full attention, making them less compatible with fine-tuning pre-trained large language models \cite{ma2024megalodon}. Another active direction involves extending the context window via continual pre-training and supervised fine-tuning on longer sequences \cite{xiong2024effective,pengyarn,chen2023extending,fu2024data,bai2024longalign}. While these methods typically requires higher computational costs, they generally achieve superior performance on a variety of long-context tasks. More recently, \cite{jin2025search} combine policy network with multi-turn search engine calling have shown significant success. In this work, we focus on long-context generation task by designing a targeted reward signal.

\subsection{Improving LLM with AI Feedback}
Reinforcement learning from human feedback (RLHF) is crucial for aligning LLMs with human values, enables them to pursue diverse goals by learning from human feedback~\cite{ouyang2022training,yuan2023rrhf,rafailov2023direct,song2024preference}. However, collecting high-quality pairwise human preference data is both expensive and time-consuming~\cite{bai2024longalign,  wu2024meta}. To address this, synthetic preference data generated by LLMs presents a promising alternative, offering scalability at a significantly lower cost. Following this direction,~\cite{bai2024longalign} first introduced LLM-generated critiques for evaluating whether model outputs are harmful, using human-annotated harmful prompts as a reference. \cite{dubois2023alpacafarm} further leveraged API-based LLMs to select preferred model responses, reducing human involvement. Reinforcement Learning with AI Feedback (RLAIF) \cite{lee2023rlaif} extends this idea by using another LLM as a verifier to approximate human judgment. Building on this, \cite{yang2024rlcd} later found that using better prompts (self-improve) that direct harmful or harmless responses can surpass RLAIF. More recently, \cite{yuan2401self} demonstrated that combining iterative fine-tuning with high-quality prompts generated via in-context learning can yield surprisingly strong performance. 

\begin{algorithm}[t!]
\caption{Synthetic preference alignment pipeline}
\label{alg:synthetic_preference}
\KwIn{Reference model $\mathcal{G}_{ref}$, preference dataset $\mathcal{D}$, iterations $W$}
\For{$w = 1, 2, \ldots, W$}{
    $\mathcal{D}_w \leftarrow$ $w$-th partition of $\mathcal{D}$\;
    \For{$m \in \mathcal{D}_w$}{
        Generate response $r$ using $\mathcal{G}_{ref}(m)$\;
    }
    $\{y, y_w, y_l\} \leftarrow$ Rank responses by preference\;
    $\mathcal{D}_w \leftarrow \{(m, y_w, y_l) \mid m \in \mathcal{D}_w\}$\;
    $\mathcal{G}_{\theta_t} \leftarrow \arg\min_\theta \mathcal{L}(\theta)$ as defined in Equation \ref{eq:MLE}\;
    where $r_\theta^*$ is given by Equation \ref{eq:dpo}\;
    $\mathcal{G}_{ref} \leftarrow \mathcal{G}_{\theta_t}$\;
}
\end{algorithm}

\subsection{LLM-as-a-Judge}
Before the era of LLMs, striking a balance between comprehensive and scalable evaluation remained a long-standing challenge \cite{gu2024survey,wang2024self}. Subjective methods such as expert-driven assessments \cite{gao2023human,shi2024judging} have long been considered the gold standard due to their ability to provide holistic reasoning and fine-grained contextual understanding. These approaches are costly, difficult to scale, and often suffer from inconsistency. In contrast, objective evaluation methods, such as automatic metrics offer strong scalability and consistency \cite{papineni2002bleu,lin2004rouge}. These metrics rely heavily on surface-level lexical overlaps, making them difficult to evaluate outputs that require deeper semantic understanding \cite{schluter2017limits}. The “LLM-as-a-Judge” paradigm has emerged as a promising alternative that combines the advantages of both paradigms: the contextual understanding of human evaluation and the scalability of automated metrics \cite{dubois2023alpacafarm,fernandes2023devil,bai2023benchmarking}. Studies have also used “LLM-as-a-Judge” to train reward models and curate preference data \cite{lee2023rlaif,chenalpagasus,liself}. However, previous methods mostly utilizing provide general assessments as reward signal, which often lack accuracy and specificity. 

\label{Related Work}

\begin{figure*}
    \centering
    \includegraphics[width=1\linewidth]{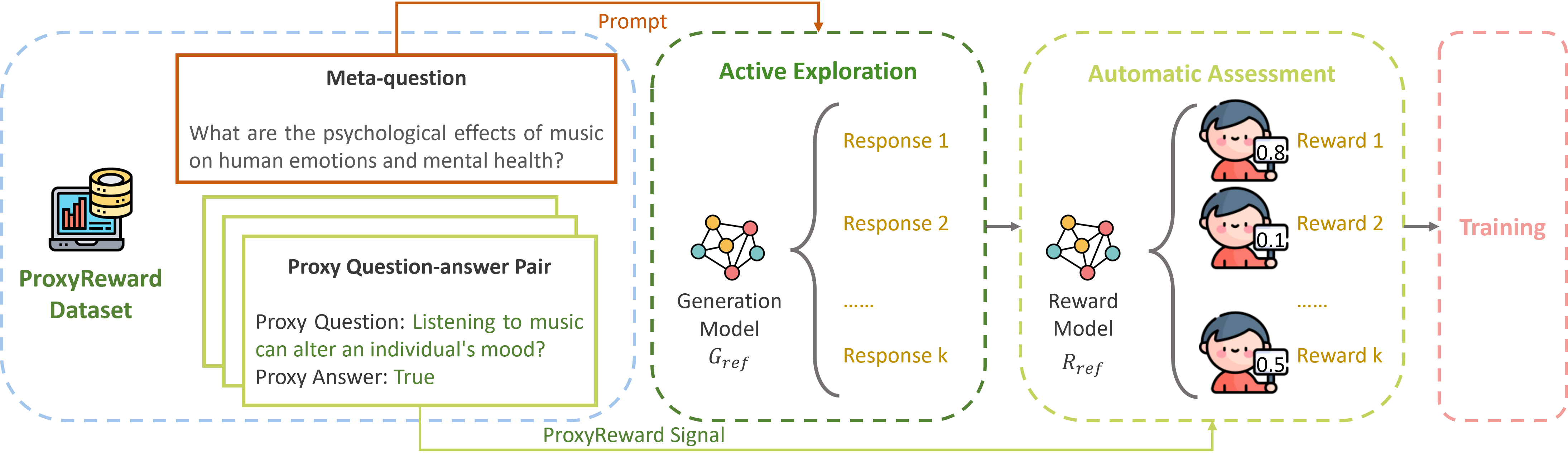}
    \caption{\ourmethod overview. Our framework operates in three stages: First, \ourmethod automatically generates a training dataset comprising meta-questions and their corresponding proxy question-answer pairs. Second, a generation model actively explores $k$ diverse responses based on these meta-questions. Third, a reward model automatically assesses these responses using the corresponding proxy question-answer pairs to generate reward signals, which are subsequently utilized during model training.}
    \label{fig:methodology}
\end{figure*}

\section{Preliminaries}

\label{Preliminaries}

\subsection{Preference Alignment}
\label{Preference Alignment}
% Preference alignment is a core process in LLMs training which aims to adapt language model's behavior to reflect human or synthetic preferences. 
Let $\mathcal{D} = \{(x, y^{+}, y^{-})\}$ denote a dataset of preferences, where $x$ is an input prompt, $y^{+}, y^{-}$ are the responses labeled as preferred and dis-preferred, respectively. The purpose of preference alignment is to designing a policy $\pi$ that maps prompts to responses, maximizing a reward that reflects human preferences using the Bradley–Terry (BT) model:

\begin{equation}
    p(y_{1} \succ y_{2}|x) = \sigma\left(r^*(x, y_{1}) - r^*(x, y_{2})\right),
\end{equation}
where $r^*(x, y_{1})$ represents the oracle reward of a response given a prompt, and
$\sigma(z) = \{1+exp(-z)\}^{-1}$ is the sigmoid function, mapping differences in rewards to probabilities. A parameterized reward model $r_{\theta}$ is estimated by solving a maximum likelihood estimation (MLE) objective:

\begin{equation}
    \mathcal{L}(\theta) = -\mathbb{E}_{(x,y_{w},y_{l}) \sim \mathcal{D}} 
    [\log \sigma (r_{\theta}^*(x, y_{w}) - r_{\theta}^*(x, y_{l})],
    \label{eq:MLE}
\end{equation}
where $y_{w}, y_{l}$ are preferred and dis-preferred sample respectively. The direct preference optimization (DPO) \cite{rafailov2023direct} we used in this work chose the reward as:

\begin{equation}
    r_\theta(x, y) = \beta \log \frac{\pi_\theta(y \mid x)}{\pi_{\text{ref}}(y \mid x)},
    \label{eq:dpo}
\end{equation}
to directly optimize the policy $\pi_{\theta}$ using the loss $L(\theta)$ in Equation \ref{eq:MLE}, with the reward function $r$ specified by Equation \ref{eq:dpo}. As the reward is implicitly defined by the policy itself, the objective becomes fully dependent on $\theta$ eliminating the need for a separately trained reward model. Consequently, this reformulation significantly improves the computational efficiency of the alignment process.

% The learned reward function is then used to provide feedback to the language model. The optimization problem is formalized as:

% \begin{equation}
%     \max_{\pi_\phi}\mathbb{E}_{x \sim \mathcal{D}, \, y \sim \pi_\phi(y \mid x)} [r(x, y)]
%     - \beta\mathrm{D}_{\mathrm{KL}} [\pi_\phi \, \| \, \pi_{\text{ref}}],
%     \label{eq:Policy Optimization.}
% \end{equation}

% where the reward term $\mathbb{E}_{x \sim \mathcal{D}, \, y \sim \pi_\phi(y \mid x)}[r(x, y)]$ encourages the policy to generate high-quality responses and the regularization term $\mathrm{D}_{\mathrm{KL}} [\pi_\phi \, \| \, \pi_{\text{ref}}]$ penalizes deviations from the reference policy $\pi_{\text{ref}}$. The $\pi$ here is defined as $\pi(y|x)$.

% \begin{equation}
%     \mathcal{L}_{\text{DPO}}(\mathcal{G}; \mathcal{D}) = -\mathbb{E}_{(m, r^{+}, r^{-}) \sim \mathcal{D}} 
%     \left[ 
%         \log \sigma 
%         \left( 
%             \beta \log \frac{\mathcal{G}(r^{+} \mid m)}{\mathcal{G}_{\text{ref}}(r^{+} \mid m)} 
%             - \beta \log \frac{\mathcal{G}(r^{-} \mid m)}{\mathcal{G}_{\text{ref}}(r^{-} \mid m)} 
%         \right)
%     \right].
% \end{equation}

% \begin{equation}
%     \mathcal{L}_{\text{DPO}}(\mathcal{G}; \mathcal{D}) = -\mathbb{E}_{(m, r^{+}, r^{-}) \sim \mathcal{D}} 
%     \left[ 
%         \log \sigma 
%         \left( 
%             \beta \log \frac{\mathcal{G}(r^{+} \mid m)}{\mathcal{G}_{\text{ref}}(r^{+} \mid m)} 
%             - \beta \log \frac{\mathcal{G}(r^{-} \mid m)}{\mathcal{G}_{\text{ref}}(r^{-} \mid m)} 
%         \right)
%     \right].
% \end{equation}

\subsection{Synthetic Preference Alignment Pipeline}
\label{Pipeline}
% Synthetic preference alignment is a scalable alternative to Reinforcement Learning from Human Feedback (RLHF), where human-labeled preference data are replaced or supplemented by synthetic feedback generated from LLMs.

Given the generation policy \(\mathcal{G}\) parameterized by $\theta$ and an LLM-based reward model \(\mathcal{R}\). As shown in Table~\ref{alg:synthetic_preference}, the synthetic preference alignment pipeline typically consists of the following stages:

\noindent \textbf{Response Generation.} Given a dataset of prompts $\mathcal{X} = \{x_{1}, \cdots, x_{n}\}$, the policy \(\mathcal{G_{\theta}}\) generate a set of responses $\{y^{1}_{i}, y^{1}_{i}, \cdots\}$ which are intended to cover diverse output patterns for each prompt $x_{i}$.

\noindent \textbf{AI-based Reward Assignment.} For each response $y^{j}_{i}$, reward score $r(x_{i},y^{j}_{i})$ is calculated by reward model \(\mathcal{R}\), which acts as an automatic evaluator.

\noindent \textbf{Policy Optimization.} The policy \(\mathcal{G_{\theta}}\) is then fine-tuned using the synthetic reward signal. DPO are commonly used to align the policy with the feedback provided by reward model \(\mathcal{R}\).

% This pipeline enables scalable and cost-efficient alignment of LLMs by automating the feedback collection process, leveraging strong LLMs as high-quality evaluators in place of human annotators.

% Requires: \usepackage{amsmath}
% \begin{equation}
%     \mathcal{L}_{\text{DPO}}(\pi; \mathcal{D}) = -\mathbb{E}_{(x, y_w, y_l) \sim \mathcal{D}} 
%     \left[ 
%         \log \sigma 
%         \left( 
%             \beta \log \frac{\pi(y_w \mid x)}{\pi_{\text{ref}}(y_w \mid x)} 
%             - \beta \log \frac{\pi(y_l \mid x)}{\pi_{\text{ref}}(y_l \mid x)} 
%         \right)
%     \right].
% \end{equation}

\section{Methodology}
\label{Methodology}
In this section, we introduce \ourmethod, a novel reinforcement learning (RL)-based framework for effective long-context generation. As shown in Figure~\ref{fig:methodology}, our approach comprises three key components: First, an automatically constructed \textbf{ProxyReward Dataset} (Section~\ref{Scoring Dataset Collection}) that utilizes LLMs to generate diverse meta-questions with corresponding proxy question-answer pairs, avoiding fixed patterns; Second, an \textbf{Active Exploration} mechanism (Section~\ref{Active Exploration}) that generates various long-form contents; Third, an \textbf{Automatic Assessment} system (Section~\ref{Automatic Assessment}) that provides targeted reward signals to guide the optimization process. This iterative process enables the model to improve by exploring, evaluating, and incorporating superior response patterns across multiple iterations. By combining automated data construction with RL-based optimization, \ourmethod efficiently addresses the challenges of \task task.

\subsection{\ourmethod Dataset Collection}
\label{Scoring Dataset Collection}

One challenge of \task is that responses do not follow a fixed pattern, making it difficult to establish standard answers when constructing the training dataset. To address this issue, we designed a \ourmethod dataset to facilitate training data generation. This dataset consists of meta-questions set \(M=\{m_{1}, m_{2}, ..., m_{n}\}\), where each meta-question requires lengthy context to answer thoroughly. For each meta-question \(m_{i}\), we develop a corresponding list of proxy question-answer pairs represented as \(P_{i}=\{(q_{i1}, a_{i1}), (q_{i2}, a_{i2}),  ..., (q_{il}, a_{il})\}\). The main idea is to transform subjective expert evaluations of long-form text quality into objective reading comprehension questions that can be automatically assessed by LLMs. 

These proxy question-answer pairs \((q, a)\) are directly related to information covered by the meta-questions \(M\). The proxy questions resemble reading comprehension questions \(Q\), with each one designed to assess a key information point that should be included in the long-form response text \(t\). All Proxy answers are formulated as boolean values. We utilize these proxy question-answer pairs \((q, a)\) as an objective checklist to quantify the information content of long-form responses. We ensure that most proxy questions have "True" as their expected answer, as "False" answers would indicate that the information in the proxy-question \(q\) is not sufficiently relevant to the meta-question \(m\).

% We posit that if an LLM-generated response \(t\) enables an LLM to correctly answer all Proxy questions, then \(t\) has achieved optimal quality. Therefore, the score for \(t\) is calculated as:

% \begin{equation}
%     S_{t_{ik}}=\frac{\sum_{j=1}^{l}\mathcal{F}(pa_{ij}, ra_{ij})}{l},
% \label{eq:score}
% \end{equation}
% \begin{equation}
%     \mathcal{F}(pa_{ij}, ra_{ij}) = \begin{cases} 
%     1 & \text{if } pa_{ij} = ra_{ij} \\
%     0 & \text{if } pa_{ij} \neq ra_{ij} 
%     \end{cases},
% \label{eq:equation}
% \end{equation}
% where \(t_{ik}\) represents the \(k\)th response of \(t_{i}\), \(pa_{ij}\) represents the predicted answer for \(q_{ij}\), and \(ra_{ij}\) represents the reference answer for \(q_{ij}\).

\begin{figure}[t!]
    \centering
    \includegraphics[width=1\linewidth]{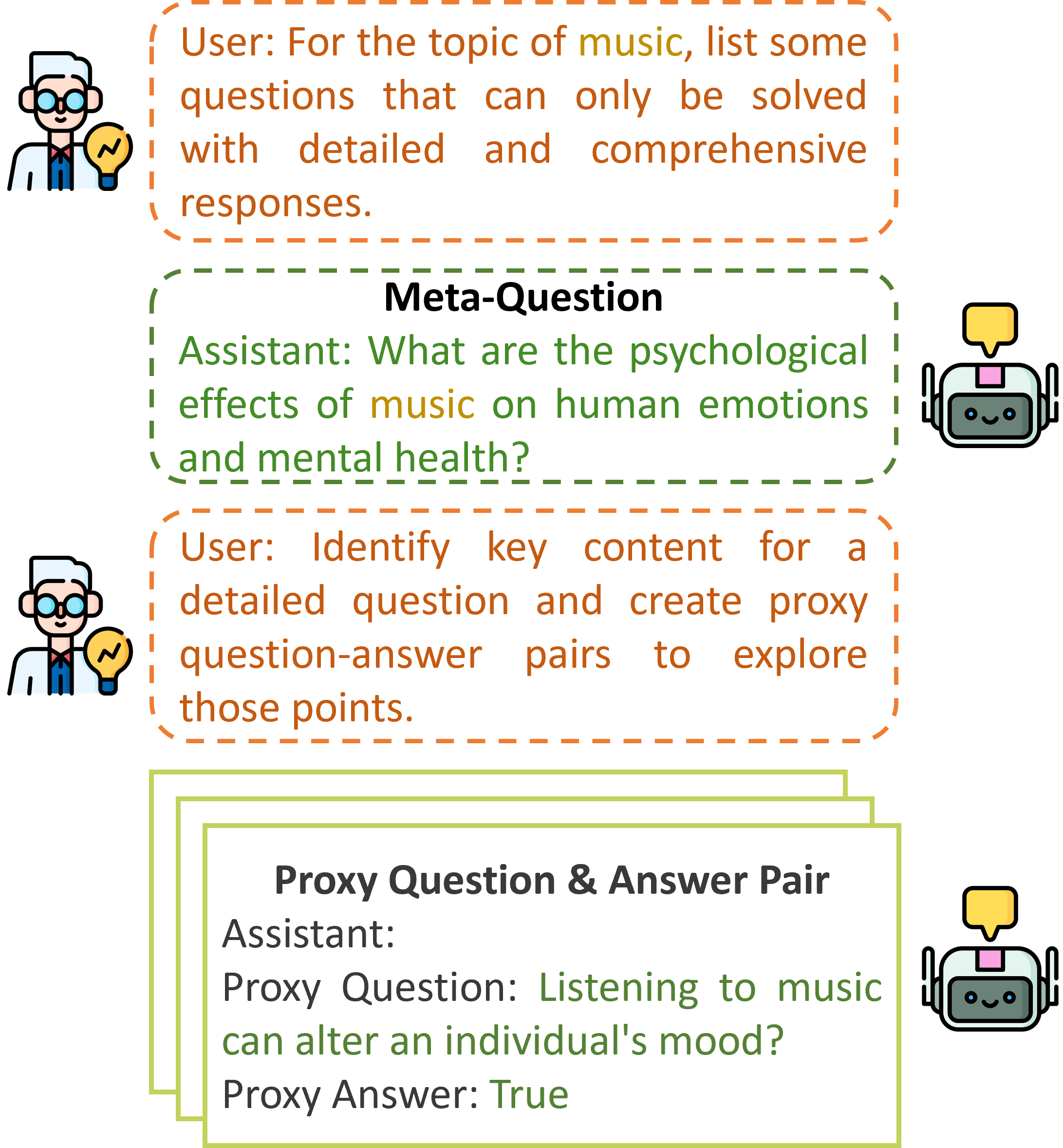}
    \caption{The pipeline of meta-questions and Proxy Question-Answer pairs generation in \ourmethod.}
    \label{fig:generation}
\end{figure}

The construction method for the \ourmethod Dataset is straightforward. Instead of traditional manual annotation, we use a simple prompt to automatically generate data through LLM API calls. We begin by manually selecting 40 domains that typically require long-form answers, including Computer Science, Technology, History, Game, Policy, and others. As shown in Figure~\ref{fig:generation}, the automatic data generation process using the LLM API consists of two key steps: First, for each domain, we prompted the LLM API to generate 9,271 meta-questions that require long-form contextual answers. Second, for each meta-question, we instructed the LLM API to generate approximately 15 boolean proxy question-answer pairs. The prompts details are shown in Appendix~\ref{sec:Prompts}. Detailed specifications of the scoring dataset are presented in Table~\ref{Dataset Details}.

\subsection{Active Exploration}
\label{Active Exploration}
A significant challenge in \task is the scarcity of large-scale human-annotated preference data exhibiting diverse patterns. To address this limitation, we implement Active Exploration, leveraging two distinct language models: a generation model \(\mathcal{G}\) and a reward model \(\mathbb{R}\). This approach proceeds in two phases. First, we extract a meta-question m from the \ourmethod Dataset and input it to our generation LLM, which produces k different responses \(T_i = {r_1, r_2, ..., r_k}\). Second, we feed both the QA list from the Scoring Dataset and the k responses \(T_i\) into a higher-performing reward LLM. This reward model comprehends all responses \(t_{ik}\) and answers all questions \(q_{ij}\), outputting an answer list \(A_{ik} = {a_{i1}, a_{i2}, ..., a_{ik}}\). Through this process, we efficiently create a large-scale training dataset for \task that contains diverse patterns and high-quality annotations.

\begin{table}[t!]
    \centering
    \caption{Statistics of the \ourmethod Dataset.}
    \begin{tabular}{cc}
    \hline
    \textbf{Dataset} & \textbf{ProxyReward} \\ \hline
    Domain           & 40                    \\
    Meta-Question    & 9,271                 \\
    Proxy Q\&A       & 156,506               \\
    Metric           & Proxy Accuracy        \\ \hline
    \end{tabular}
    \label{Dataset Details}
\end{table}

\begin{table*}[t!]
    \centering
    \caption{Comparison of Accuracy (\%) of ProxyQA across different settings.}
    \begin{tabular}{lcccc}
    \toprule
    \textbf{Model} & \textbf{Base} & \textbf{LLM-as-a-Judge} & \textbf{\ourmethod} & \textbf{\ourmethod} \\
     & & & \textbf{(Iter 1)} & \textbf{(Iter 2)} \\
    \midrule
    % Qwen models
    Qwen2.5-1.5B-Instruct & 18.30 & 12.88 & \textbf{22.10} & 21.54 \\
    Qwen2.5-3B-Instruct & 27.00 & 16.69 & 27.58 & \textbf{28.54} \\
    Qwen2.5-7B-Instruct & 32.37 & 25.73 & 33.23 & \textbf{35.07} \\
    \midrule
    % LLaMA models
    Llama-3.2-1B-Instruct & 18.49 & 11.84 & 19.97 & \textbf{20.08} \\
    Llama-3.2-3B-Instruct & 25.16 & 21.16 & 26.96 & \textbf{28.59} \\
    Llama-3.1-8B-Instruct & 25.02 & 23.43 & 29.38 & \textbf{30.11} \\
    \bottomrule
    \end{tabular}
    \label{tbl: Main results}
\end{table*}

% \subsection{Reward Signal}
% \label{Reward Signal}
\subsection{Automatic Assessment}
\label{Automatic Assessment}
To incorporate expert preferences while minimizing annotation costs, we introduce automatic assessment as the reward signal.   Specifically, we hypothesize that the quality of a long-form response \(r_i\) is optimal when it enables a language model to accurately answer all associated proxy questions \(P_i = \{(\hat{q}_{ij}, \hat{a}_{ij})\}_{j=1}^l\). A response \(r_i\) is deemed thorough and informative if it facilitates correct answers to all proxy questions, the prompt detail as shown in Appendix~\ref{sec:Prompts}. Conversely, if \(r_i\) fails to correctly answer a significant portion of the proxy questions, it lacks the necessary information and coherence required for a high-quality long-form response. Building on this intuition, we design the reward signal as a scoring function $\mathcal{S}(r_i)$ that quantifies the informativeness of \(r_i\). The reward function is defined as:
\begin{equation}
    \mathcal{S}(r_i) = \frac{\sum_{j=1}^{l}\mathcal{F}(a'_{ij}, \hat{a}_{ij})}{l},
\label{eq:score}
\end{equation}
where \(\mathcal{F}(a'_{ij}, \hat{a}_{ij})\) is a binary function that compares the predicted answer \(a'_{ij}\) for proxy question \(\hat{q}_{ij}\) with the reference answer \(\hat{a}_{ij}\):
\begin{equation}
    \mathcal{F}(a'_{ij}, \hat{a}_{ij}) = 
    \begin{cases} 
    1, & \text{if } a'_{ij} = \hat{a}_{ij}, \\
    0, & \text{if } a'_{ij} \neq \hat{a}_{ij}.
    \end{cases}
\label{eq:comparison}
\end{equation}
Here, \(a'_{ij} = R_{ref}(t_i, \hat{q}_{ij})\) represents the predicted answer generated by the reward model \(R_{ref}\) when evaluating the response \(r_i\) against the proxy question \(\hat{q}_{ij}\). The reference answer \(\hat{a}_{ij}\) serves as the ground truth for \(\hat{q}_{ij}\).

The function \(\mathcal{S}(r_i)\) measures the proportion of correctly answered proxy questions, replacing rigid evaluation metrics for long-form text quality with a more adaptable framework. This flexible approach enables the reward signal to effectively steer the reinforcement learning process, fostering the generation of responses that are not only informative but also deeply aligned with the context of the tasks.

\section{Experiment}
\label{Experiment}

\subsection{Setup}
\noindent \textbf{Baseline Models} For trainable baselines, we use Llama (including Llama-7B~\cite{touvron2023llama}, Llama-2-7B~\cite{touvron2023llama}, Llama-2-13b~\cite{touvron2023llama}, Llama-3.2-1B-Instruct~\cite{grattafiori2024llama}, Llama-3.2-3B-Instruct~\cite{grattafiori2024llama} and Llama-3.1-8B-Instruct~\cite{grattafiori2024llama}), and Qwen Instruct (including Qwen2.5-1.5B-Instruct~\cite{team5qwen2}, Qwen2.5-3B-Instruct~\cite{team5qwen2}, and Qwen2.5-7B-Instruct~\cite{team5qwen2}) as backbone models. Training-free baselines include GPT (including GPT-3.5-Turbo, GPT-4, GPT-4-Turbo, GPT-4o-mini~\cite{openai2023gpt4omini} and GPT-4o~\cite{openai2023gpt4o}), DeepSeek (including DeepSeek-V3~\cite{liu2024deepseek} and DeepSeek-R1~\cite{guo2025deepseek}), Bing~\cite{bing2023}. We also compare out method with LLM-as-a-Judge~\cite{gu2024survey}.  The LLM-as-a-Judge prompt is shown in Appendix~\ref{sec:Prompts}.

\noindent \textbf{Evaluation and Metrics} We assess performance using the ProxyQA benchmark~\cite{tan2024proxyqa}, an innovative dataset designed for evaluating long-text generation. ProxyQA consists of in-depth, human-curated meta-questions across various domains, each accompanied by specific proxy questions and pre-annotated answers.

\noindent \textbf{Data Selection} The quality of responses automatically generated by the generation model varies considerably, which significantly impacts training outcomes. To address this challenge, we implement data selection based on scores. First, we recognize that meta-questions vary in difficulty—for simpler questions, consistently produces high-scoring responses, while for more challenging questions, the scores of generated responses show greater variance. Consequently, we prioritize meta-questions that exhibit higher variance in scores for the same response. This approach prevents our dataset from being dominated by simple questions, which would reduce training efficiency. Second, to avoid model optimization direction contrast, we filter cases where a single meta-question has multiple preference indicators. From the remaining data, we select the highest and lowest scoring responses to create preference pairs, which serve as reference responses for training model. The selection details are in Table~\ref{tbl:selection}.

\noindent \textbf{Model Settings} We utilize the GPT-4o-mini API to construct the \ourmethod dataset. All methods are trained using consistent hyperparameters across the board. Training is conducted on \(4 \times L40\) GPUs. The learning rate for DPO is set at 5e-7, with a batch size of 2 and a gradient accumulation step of 8. The maximum completion length is 2048 tokens, and we train for 5 epochs. To enhance the diversity of the outputs generated by the LLMs, we implement a temperature setting of 0.8 for the trainable models. For the reward model, we employ GPT-4o-mini~\cite{openai2023gpt4omini} to compute training rewards. For evaluation purposes, we utilize GPT-4o~\cite{openai2023gpt4o} to determine the ProxyQA score.

% \noindent \textbf{Model Settings} We use GPT-4o-mini api to build \ourmethod Dataset. All methods are trained with uniform hyperparameters. We train on 4\(\times\)L40 GPUs. For learning rate, DPO use 5e-7. Batch size is 2. Gradient accumulation step is 8. Max completion length is 2048. The number of train epochs is 5. To increase the diversity of LLMs' output, a temperature setting of 0.8 is implemented for trainable models. For reward model, we use GPT-4o-mini~\cite{openai2023gpt4omini} to calculate training reward. For evaluation, we use GPT-4o~\cite{openai2023gpt4o} to calculate ProxyQA score.

\begin{table}[t!]
    \centering
    \caption{Comparison of ProxyQA Accuracy (\%) between \ourmethod and closed-source LLMs. We use \textit{Iter 2} \ourmethod results for both Qwen and Llama.}
    \begin{tabular}{lc}
    \toprule
    \textbf{Model} & \textbf{ACC} \\
    \midrule
    % Qwen models
    Qwen2.5-7B-Instruct & \textbf{35.07} \\
    Llama-3.1-8B-Instruct & 30.11 \\
    \midrule
    GPT-3.5-Turbo & 23.94 \\
    GPT-4-0613 & 27.19 \\
    GPT-4-Turbo & 33.94 \\
    GPT-4o-mini & 37.57 \\
    GPT-4o & 44.98 \\
    \midrule
    DeepSeek-V3 & 42.93 \\
    DeepSeek-R1 & \textbf{48.65} \\
    \midrule
    ReAct (GPT-4) & 17.15 \\
    ReAct (GPT-4-Turbo) & 21.19 \\
    Bard (Gemini Pro) & 25.00 \\
    New Bing (Creative Mode) & 39.37 \\
    \bottomrule
    \end{tabular}
    \label{tbl:closed-source}
\end{table}

\subsection{How does \ourmethod Performance Compare to Trainable Baselines?}
\textit{Our proposed \ourmethod demonstrated substantial improvements on open-source models.} The experimental results clearly demonstrate the effectiveness of our \ourmethod method across different model architectures and parameter scales. As shown in Table~\ref{tbl: Main results}, \ourmethod consistently outperforms both the base models across all tested configurations. Notably, when applied to the Qwen2.5-7B-Instruct model, our method achieves a remarkable score of 35.07, surpassing even GPT-4-Turbo (33.94) which is a significantly larger proprietary model. This performance superiority highlights the efficiency of our approach in enhancing model capabilities without requiring the massive computational resources typically associated with larger models.

The improvement pattern is consistent across different model scales. For smaller models like Qwen2.5-1.5B-Instruct, \ourmethod boosts performance from 18.30 to 22.10, representing a 20.8\% relative improvement. Similarly, for medium-sized models such as Qwen2.5-3B-Instruct, our method increases the score from 27.00 to 28.54. The enhancement is equally evident in the Llama series, where Llama-3.1-8B-Instruct shows a substantial improvement from 25.02 to 30.11, demonstrating a 20.3\% relative gain.

% These results validate that \ourmethod provides an efficient framework for enhancing model performance across different architectures and scales, offering a viable approach to achieve state-of-the-art performance with relatively smaller models.

These results validate that \ourmethod provides an efficient framework for enhancing model performance across architectures and scales, offering a viable approach to achieve state-of-the-art performance with smaller models.

% \subsection{How does \ourmethod Performance Compare to Trainable Baselines?}
% \textit{Our proposed \ourmethod demonstrated substantial improvements on open-source models.} As shown in Table~\ref{tbl: Main results}, \ourmethod improve the performance on Llama models. Llama-3.1-8B-\ourmethod achieves a 5-point improvement (30.11) over Llama-3.1-8B-Instruct (25.02). Llama-3.2-3B-\ourmethod also outperforms Llama-3.2-3B-Instruct (from 25.16 to 28.54), shows the effectiveness of \ourmethod reward signal. Llama-3.2-1B-\ourmethod also outperforms Llama-3.2-1B-Instruct (from 18.49 to 20.08). Results show the effectiveness of \ourmethod reward signal.

% Qwen powered by \ourmethod shows similar improvements. Qwen2.5-7B-\ourmethod achieves a 3-point improvement over Qwen2.5-7B-Instruct, reaching 35.07. Qwen2.5-3B-\ourmethod achieves a 3-point improvement over Qwen2.5-3B-Instruct, reaching 35.07. Qwen2.5-1.5B-\ourmethod also outperforms Qwen2.5-1.5B-Instruct (from 18.30 to 22.10).

\begin{figure}[t!]
    \centering
    \includegraphics[width=1\linewidth]{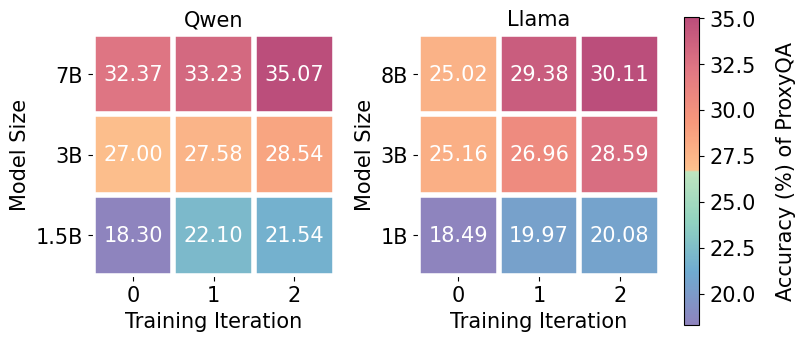}
    \caption{Effects of model size and iteration on ProxyQA score for 3×3.}
    \label{fig:iteration}
\end{figure}

\subsection{How does \ourmethod Performance Compare to LLM-as-a-Judge Performance?}
% \textit{Comparing to LLM-as-a-Judge, the \ourmethod signal offers a targeted evaluation of information comprehensiveness and accuracy for specific questions.}

\textit{Compared to the LLM-as-a-Judge approach, the \ourmethod signal offers a targeted evaluation of information comprehensiveness and accuracy for specific questions, resulting in higher performance.} To evaluate the effectiveness of our proposed \ourmethod signal, we conducted comparative experiments against the conventional LLM-as-a-Judge approach using Qwen2.5 models at 1.5B, 3B and 7B parameter scales. The results demonstrate a substantial performance advantage for \ourmethod across all model sizes.

As shown in Table~\ref{tbl: Main results}, the \ourmethod signal achieved scores of 22.10, 28.54 and 35.07 for the 1.5B, 3B and 7B models respectively, compared to 12.88, 16.69 and 25.73 for the LLM-as-a-Judge approach. This represents an absolute improvement of an increase of 9.22 points for the 1.5B model, 11.85 points for the 3B model, and 9.34 points for the 7B model. This corresponds to relative improvements of approximately 71.6\%, 71.1\%, and 36.3\%, respectively. These enhancements highlight the superior effectiveness of \ourmethod in comparison to the LLM-as-a-Judge approach.

Similar to Qwen, our method demonstrates strong performance on the Llama models. The \ourmethod signal achieved scores of 20.08, 28.59, and 30.11 for the 1B, 3B, and 8B models, respectively, while the LLM-as-a-Judge approach scored 18.49, 25.16, and 25.02 for the same models. This results in absolute improvements of 1.59 points for the 1B model, 3.43 points for the 3B model, and 5.09 points for the 8B model. Correspondingly, the relative improvements are approximately 8.6\%, 13.6\%, and 20.4\%. These results further underscore the effectiveness of \ourmethod in enhancing model performance across different sizes of Llama models.

These results confirm our hypothesis that unlike LLM-as-a-Judge approaches, which provide general assessments of text quality, the \ourmethod signal delivers more targeted evaluation of information comprehensiveness and accuracy for specific questions. This approach appears particularly advantageous when scaling to larger model sizes, indicating better alignment with the evaluation criteria required for our task.

\subsection{How does \ourmethod Performance Compare to Closed-source Baselines?}
\textit{Our method can improve small size models to exceed GPT-4-Turbo.} As shown in Table~\ref{tbl:closed-source}, Qwen2.5-7B-Instruct achieves an accuracy of 35.07\% on the ProxyQA benchmark, surpassing GPT-4-Turbo (33.94\%) despite having a substantially smaller parameter count. This is a remarkable achievement considering the vast resource difference between these models.

The performance gap is even more pronounced when comparing to GPT-3.5-Turbo (23.94\%) and GPT-4-0613 (27.19\%), with our Qwen2.5-7B-Instruct outperforming them by 11.13 and 7.88 percentage points respectively. While Llama-3.1-8B-Instruct achieves a lower accuracy of 30.11\%, it still exceeds GPT-3.5-Turbo and GPT-4-0613, demonstrating the effectiveness of our approach across different model architectures.

\begin{table}[t!]
    \centering
    \caption{Ablation study of ProxyQA Accuracy (\%) in the Qwen 2.5 series.}
    \begin{tabular}{lccc}
    \toprule
    \textbf{Model} & \textbf{1.5B} & \textbf{3B} & \textbf{7B} \\
    \midrule
    ProxyReward & \multirow{2}{*}{12.57} & \multirow{2}{*}{17.49} & \multirow{2}{*}{25.67} \\
    \footnotesize{    - Precision} & & & \\
    ProxyReward & \multirow{2}{*}{\textbf{22.10}} & \multirow{2}{*}{27.58} & \multirow{2}{*}{33.23} \\
    \footnotesize{    - Accuracy (Iter 1)} & & & \\
    ProxyReward & \multirow{2}{*}{21.54} & \multirow{2}{*}{\textbf{28.54}} & \multirow{2}{*}{\textbf{35.07}} \\
    \footnotesize{    - Accuracy (Iter 2)} & & & \\
    \bottomrule
    \end{tabular}
    \label{tbl:ablation}
\end{table}
% \vspace{-0.5em}
\section{Ablation Study}
In this section, we conduct an ablation study to systematically investigate the impact of various factors on the performance of our proposed model. We address two critical questions: first, we explore whether multiple training iterations with ProxyReward can enhance model performance Section~(\ref{sec:iteration}); second, we evaluate the effectiveness of different ProxyReward metrics to identify the most suitable one for our tasks Section~(\ref{sec:metric}).

\subsection{Can \ourmethod Multiple Training Iterations Improve Performance?}
\label{sec:iteration}
% \textit{The results reveal a consistent pattern of improvement when applying \ourmethod, particularly with multiple training iterations.} Across all tested models, multiple iterations of \ourmethod consistently yielded better performance compared to single iterations. For instance, Qwen2.5-3B-Instruct improved from 27.58 to 28.54, while Llama-3.1-8B-Instruct showed substantial gains from 29.38 to 30.11. Even the smallest models in our evaluation benefited from this iterative approach, with Qwen2.5-1.5B-Instruct and Llama-3.2-1B-Instruct showing improvements of 21.54 and 20.08, respectively.

% \textit{The results reveal a consistent pattern of improvement when applying \ourmethod, especially with multiple training iterations.} Across all tested models, multiple iterations of \ourmethod consistently yielded better performance than single iterations. For instance, Qwen2.5-3B-Instruct improves from 27.58 to 28.54, while Llama-3.1-8B-Instruct shows gains from 29.38 to 30.11. Even the smallest models benefited from this iterative approach, with Qwen2.5-1.5B-Instruct and Llama-3.2-1B-Instruct showing improvements of 21.54 and 20.08, respectively.

\textit{The results reveal a consistent pattern of improvement when applying \ourmethod, especially with multiple training iterations.} Across all tested models, multiple iterations of \ourmethod consistently outperform single iterations. For example, Qwen2.5-3B-Instruct improves from 27.58 to 28.54, while Llama-3.1-8B-Instruct increases from 29.38 to 30.11. Even the smallest models benefited, with Qwen2.5-1.5B-Instruct and Llama-3.2-1B-Instruct showing improvements of 21.54 and 20.08, respectively.

% Interestingly, the conventional LLM-as-a-Judge approach consistently underperformed compared to both baseline models and our method, suggesting potential limitations in directly applying judgment signals without the proxy mechanism we propose. This performance gap was particularly pronounced in smaller models, highlighting the scalability advantages of our approach across different model sizes.

Interestingly, the conventional LLM-as-a-Judge approach consistently underperformed compared to both baseline models and our method, suggesting limitations in directly applying judgment signals without our proposed proxy mechanism. This performance gap was especially pronounced in smaller models, highlighting the scalability advantages of our approach across model sizes.

These results collectively validate that \ourmethod not only enhances model performance but can achieve state-of-the-art results that exceed those of much larger models when applied iteratively, offering an efficient pathway to improved capabilities without the computational costs associated with scaling model size.

\begin{table*}[t!]
    \centering
    \caption{Statistical characteristics (\%) of response quality distributions across different models and reward approaches.}
    % \caption{Data Selection Details.}
    \begin{tabular}{llcccc} 
    \toprule
    \textbf{Reward}               &      \textbf{Base Model}            & \textbf{Mean} & \textbf{Variance} & \textbf{Largest Mean} & \textbf{Lowest Mean}  \\ 
    \hline
    \multirow{3}{*}{LLM-as-a-Judge} & Qwen2.5-1.5B-Instruct         & 73.00      & 0.68     & 80.98                 & 50.53                 \\
                                    & Qwen2.5-3B-Instruct           & 99.02      & 0.18     & 99.92                 & 78.88                 \\ 
                                    & Qwen2.5-7B-Instruct           &   62.42    &   1.16   &   77.58               &     39.22             \\ 
    \hline
    \multirow{3}{*}{\ourmethod \footnotesize{    - Precision}}      & Qwen2.5-1.5B-Instruct         & 82.80      & 5.71     & 99.24                 & 3.35                  \\
                                    & Qwen2.5-3B-Instruct           & 84.90      & 4.07     & 99.24                 & 6.87                  \\ 
                                    & Qwen2.5-7B-Instruct           &   84.18    &   5.75   &   99.92               &     1.25              \\ 
    \hline
    \multirow{6}{*}{\ourmethod \footnotesize{    - Accuracy}}     & Qwen2.5-1.5B-Instruct         & 37.94          & 1.52      & 63.53                  & 18.7            \\
                                    & Qwen2.5-3B-Instruct        & 49.12      & 1.34     & 70.05              & 27.77                       \\
                                    & Qwen2.5-7B-Instruct        & 50.65      & 1.47     & 72.57              & 28.19                       \\ 
        &  Llama-3.2-1B-Instruct      & 48.02      & 0.66     & 62.01              & 34.45                       \\
                                    & Llama-3.2-3B-Instruct      & 23.34      & 1.39     & 53.51              & 7.58                        \\
                                    & Llama-3.1-8B-Instruct & 30.84      & 2.13     & 62.54              & 9.52                        \\
    \bottomrule
    \end{tabular}
    \label{tbl:selection}
\end{table*}

\subsection{Which \ourmethod Metric Should We Choose?}
\label{sec:metric}
\textit{The results demonstrate that accuracy-based metrics consistently outperform precision-based metrics across all model scales.} In particular, accuracy measurements after multiple iterations show the most substantial performance gains for larger models. As shown in Table~\ref{tbl:ablation}, based on our ablation study examining different evaluation metrics across varying model sizes (1.5B, 3B, and 7B parameters), we established a systematic methodology for metric selection. When comparing the metrics directly, we observe that accuracy after the second iteration yields the best overall performance for the 7B parameter model (35.0), representing a significant improvement over precision-based evaluation (25.67). For mid-sized models (3B parameters), second-iteration accuracy also demonstrates superior performance (28.54), though with a less pronounced advantage compared to first-iteration accuracy (27.58).

Interestingly, for the smallest model (1.5B parameters), first-iteration accuracy (22.10) slightly outperforms second-iteration accuracy (21.54), suggesting that additional iterations may introduce noise rather than refinement at smaller parameter scales. This pattern indicates that optimal metric selection should be calibrated to model size, with larger models benefiting from extended iterative evaluation approaches.

\section{Conclusion}
In this paper, we addressed a critical gap in LLM research by focusing on \task. We identified two key challenges: the scarcity of gold-standard reference data for training and the inadequacy of general assessment methods as reward signals. To address these challenges, we introduced \textbf{\ourmethod}, a novel reinforcement learning (RL)-based framework that comprises two main components. First, our dataset generation methodology eliminates the need for extensive labeled data or manual effort by employing automatic creation of training examples. Second, our \ourmethod signal design provides precise evaluation of information comprehensiveness and accuracy, moving beyond the limitations of general assessment methods. Experimental results demonstrate the effectiveness of \ourmethod, with performance improvements of up to 20\% on \task tasks when applied to widely used open-source models. Notably, our approach outperforms both the LLM-as-a-Judge methodology and even GPT-4-Turbo in comparative evaluations. This work establishes a framework for enhancing LLMs' capabilities in generating coherent, informative, and contextually grounded long-form responses to complex human queries.

\section{Limitations}
While ProxyReward demonstrates significant improvements in open-ended long text generation, several limitations remain. The reliance on LLM-generated proxy question-answer pairs introduces potential biases and errors inherent to the underlying models, which may affect the objectivity and coverage of the reward signals. Additionally, the framework’s effectiveness depends on the quality and diversity of automatically generated meta-questions, which may not fully capture the complexity of real-world user queries across all domains. Finally, although ProxyReward reduces the need for manual annotation, it still requires access to high-performing LLMs for reward computation, which may pose computational and cost challenges for some practitioners.

\bibliography{custom}

\appendix

\section{Prompts}
\label{sec:Prompts}

In this part, we first present the prompts used in the two main components of \ourmethod dataset construction: meta-question and proxy question-answer pair generation.

\begin{promptbox}[Prompt for Meta-question Generation]{lightblue}
You are a data scientist. For the specified topic, please provide a list of questions that require detailed and comprehensive responses. Your tone should be formal.\\

Topic: \{TOPIC\}\\

Here is an example: \\

Input: \\

Topic: Computer Science \\

Output: \\

Data parallelism, model parallelism, and pipeline parallelism play a vital role in the training of large-scale language models. What are the representative works and frameworks among these technologies? Please introduce these technologies and frameworks in detail.
\end{promptbox}

\begin{promptbox}[Prompt for Proxy Question-answer Pair Generation]{lightblue}
You are a data scientist. Your task is generate proxy question-answer pairs baesd on given meta-question.\\

Meta-questions are defined as questions that require detailed and comprehensive responses. \\

For a given meta-question, please identify the key content necessary for formulating a detailed question and create more than 15 proxy question-answer pairs to explore these points. \\

Each proxy question should incorporate a key aspect of the meta-question. \\

The corresponding proxy answers should be one of the following: \{True, False, Not Mentioned\}, indicating the correctness and relevance of each proxy question to the meta-question.\\

Meta-question: \{META-QUESTION\}\\

Here is an example: \\

Input: \\

Meta-question: Contrastive learning has greatly promoted the progress of the learning of sentence embeddings. Please introduce some effective contrastive learning methods in sentence embedding. \\

Output: \\

1. \textbf{Question}: The hierarchical sampling strategy first selects a subset of negative samples based on their relevance to positive samples, then randomly samples from this subset to form hard negatives. \\
\textbf{Answer}: True\\

2. \textbf{Question}: Given a sentence, EDA (Easy Data Augmentation) randomly chooses and applies one of four simple operations: Synonym replacement (SR), Random insertion (RI), Random swap (RS), and Random deletion (RD).\\
\textbf{Answer}: True\\

3. \textbf{Question}: SBERT (Sentence-BERT) relies on siamese and triplet network architectures to learn sentence embeddings such that the sentence similarity can be estimated by cosine similarity between pairs of embeddings.\\
\textbf{Answer}: True\\

4. \textbf{Question}: BERT-flow was proposed to transform the embedding into a smooth and isotropic Gaussian distribution via normalizing flows.\\
\textbf{Answer}: True\\

5. \textbf{Question}: IS-BERT (Info-Sentence BERT) adopts a self-supervised learning objective based on mutual information maximization to learn good sentence embeddings in an unsupervised manner.\\
\textbf{Answer}: True\\
\end{promptbox}

Moreover, we present the prompt of \ourmethod automatic assessment.
\begin{promptbox}[Prompt for \ourmethod Signal]{lightblue}
Read the provided document and determine whether the question or statement below is "True", "False" or "Not mentioned". \\

Use only the information in the text to make your decision. Do not rely on prior knowledge or information outside of the given text. \\

If the text does not provide enough information to make a decision, respond with "Not mentioned". \\

Your are required to explain how you answer the question, and then select the final answer from "True", "False" and "Not Mentioned". \\

Document: \{DOCUMENT\}\\

Question: \{QUESTION\}
\end{promptbox}

Finally, we show the prompts used in LLM-as-a-Judge baseline.

\begin{promptbox}[Prompt for LLM-as-a-Judge Reward Signal]{lightblue}
Evaluate the quality of the given response to the question. \\

Rate the response on four dimensions: accuracy, coherence, factuality, and comprehensiveness. Use a scale from 1 (worst) to 10 (best). \\

\begin{itemize}
    \item Accuracy: Assess how well the response addresses the question and provides correct information.
    \item Coherence: Evaluate the clarity and logical flow of the response.
    \item Factuality: Check for the presence of verifiable facts and data.
    \item Comprehensiveness: Determine if the response covers all relevant aspects of the question.
\end{itemize}

Be strict in your evaluation, and aim to use the full scale. Consider the following criteria for scoring: \\

\begin{itemize}
    \item A score of 1-3 indicates major flaws in multiple dimensions.
    \item A score of 4-6 indicates moderate issues or inconsistencies.
    \item A score of 7-8 reflects generally good quality with minor flaws.
    \item A score of 9-10 is reserved for exemplary responses that excel in all dimensions.
\end{itemize}

Question: \{META-QUESTION\} \\

Response: \{DOCUMENT\}
\end{promptbox}

% \section{Further Analysis}
% \label{Further Analysis}

% \subsection{How does Data Selection affect Performance?}

% \textit{Our experiments demonstrate that data selection strategy significantly impacts \ourmethod on model performance, with the <largest,lowest> approach yielding the best results.} As shown in Table~\ref{tbl:selection}, selecting the largest examples from the source language and the lowest examples from the target language achieved a score of 19.97, outperforming both the baseline <lowest,largest> strategy (19.32) and the random selection approach (19.49). This suggests that prioritizing larger, more complex examples from the source language while focusing on simpler structures in the target language creates a more effective learning dynamic for cross-lingual transfer. The performance difference of 0.65 points between the best and worst strategies highlights the importance of thoughtful data selection in optimizing NLP models for cross-lingual tasks.
% \subsection{How Does Model Size Effect on Performance?}

% \begin{figure}
%     \centering
%     \includegraphics[width=1\linewidth]{images/selection.png}
%     \caption{Comparison of data selection strategies for \ourmethod training.}
%     \label{fig:Data Selection}
% \end{figure}

\end{document}